%% file: main.tex
\documentclass[onecolumn, crcready]{iosart2x}

\usepackage{graphicx}
\graphicspath{ {./graphics/} }
\usepackage{tikz}
\usepackage{subcaption}

\pubyear{0000}
\volume{0}
\firstpage{1}
\lastpage{1}

\begin{document}
\begin{frontmatter}

\title{Perspectives on the System-level Design of a Safe Autonomous Driving Stack}
\runtitle{System-level Design of a Safe AD Stack}

\begin{aug}
    \author[A]{\inits{M.}\fnms{Majd} \snm{Hawasly}\ead[label=e1]{first.last@five.ai}%
    \thanks{Corresponding author. \printead{e1}.}}
    \author[A]{\fnms{Jonathan} \snm{Sadeghi}}
    \author[A]{\fnms{Morris} \snm{Antonello}}
    \author[A,B]{\fnms{Stefano} \snm{V. Albrecht}}
    \author[A]{\fnms{John} \snm{Redford}}
    \author[A,B]{\fnms{Subramanian} \snm{Ramamoorthy}}
    \address[A]{\orgname{Five AI Ltd.}, \cny{United Kingdom}\printead[presep={\ ---\ }]{e1}}
    \address[B]{School of Informatics, \orgname{University of Edinburgh}, Edinburgh, \cny{United Kingdom}}
\end{aug}

\begin{abstract}
    Achieving safe and robust autonomy is the key bottleneck on the path towards broader adoption of autonomous vehicles technology. This motivates going beyond extrinsic metrics such as miles between disengagement, and calls for approaches that embody safety by design. In this paper, we address some aspects of this challenge, with emphasis on issues of motion planning and prediction. We do this through description of novel approaches taken to solving selected sub-problems within an autonomous driving stack, in the process introducing the design philosophy being adopted within Five. This includes safe-by-design planning, interpretable as well as verifiable prediction, and modelling of perception errors to enable effective sim-to-real and real-to-sim transfer within the testing pipeline of a realistic autonomous system.
\end{abstract}


\end{frontmatter}

\begin{figure}[h]
\centering
\hspace{-40px}
\begin{minipage}{.4
\textwidth} 
\includegraphics[height=150px]{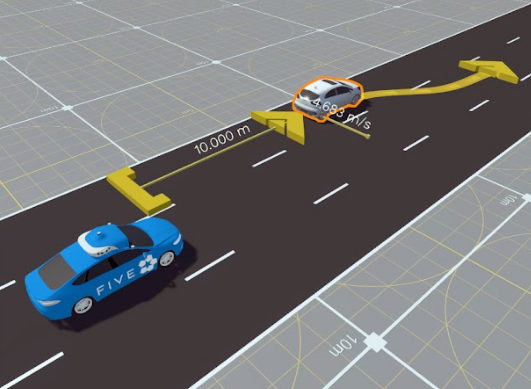}
\captionof{figure}{A simulation of an  automated lane keeping system (ALKS) scenario.}
\label{fig:alks}
\end{minipage}
\hspace{30px}
\begin{minipage}{.4\textwidth} 
\includegraphics[height=150px]{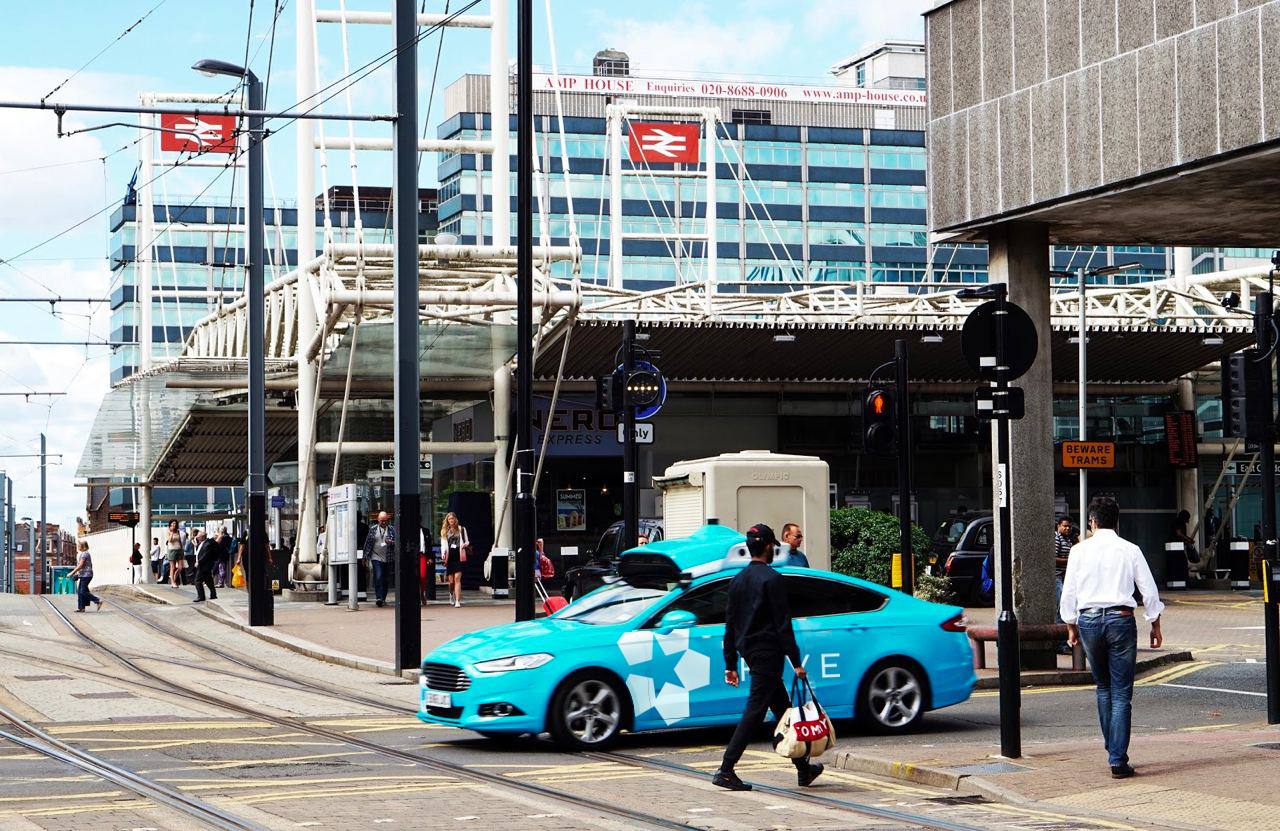}
\caption{Five's physical self-driving platform traversing a complex London intersection.}
\label{fig:london_intersection}
\end{minipage}%
\end{figure}

\section{Introduction}
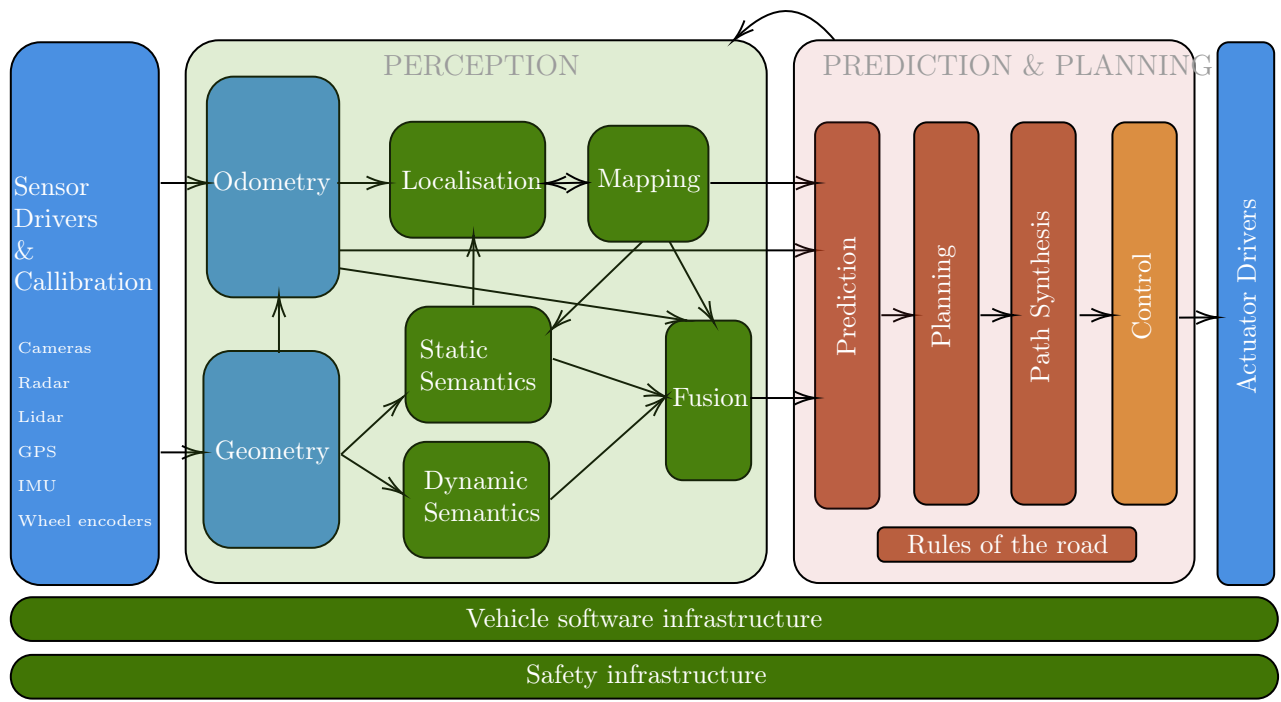
\begin{figure}
    \centering
    \input{stack.tikz}
    \caption{
        A typical functional block diagram of a classic autonomous driving software stack.
    }
    \label{fig:func_block_1}
\end{figure}

The technology stack for autonomous driving requires bringing together a number of different capabilities (Figure~\ref{fig:func_block_1}) to achieve an end goal of high-reliability system-level behaviour. As the technology has matured over the decades since the earliest experimental prototypes, going back to the middle of the twentieth century if not earlier, the emphasis has shifted in terms of which set of capabilities present the most difficulty. This also reflects that the maturing of some components have caused other capabilities to become necessary. As an example, with localisation and object detection modules becoming more reliable, it becomes possible to deploy systems in environments where {\textit{interactive}} decision making becomes essential.

The purpose of this article is to give a selective overview of activities within Five, focusing attention on elements of  safe-by-design motion planning and prediction, in particular multi-agent interactive issues, together with demonstrating safety using large scale simulation of relevant driving scenarios with realistic models of perception errors.  In the process, we touch upon the dependence of these capabilities on core functionality for perception, sensor fusion and localisation, but in the interest of space we will not discuss those components in detail.

\subsection{Driving as interactive decision making in a multi-agent environment}

Driving is an intrinsically multi-agent activity, even though the focus of attention is often on the task of the single agent under one's control. Even a seemingly standalone task, such as automated lane-keeping\footnote{UN Regulation No. 157 - Automated Lane Keeping Systems (ALKS), Doc. symbol: E/ECE/TRANS/505/Rev.3/Add.156, https://unece.org/transport/documents/2021/03/standards/un-regulation-no-157-automated-lane-keeping-systems-alks} (Figure~\ref{fig:alks}) or intersection management in
large roads (Figure~\ref{fig:london_intersection}), involves requirements that are defined in terms of responses to other agents and already require basic models of other agents' behaviours. More sophisticated urban driving scenarios (e.g. merging in high speed roads) involve not only much more elaborate models of driving behaviours, but also the need for tightly coupled interactions between one's own vehicle (often termed \textit{``ego-vehicle''}) and other road users~\cite{schwarting2018planning, 8667866}.

\subsection{Specifications for safe driving}
Ideas from control theory have a played a central role in the design of automated agents in transportation systems~\cite{hedrick1994control}, motivated by the need for robust and stable operation in the face of uncertainty~\cite{zhou1998essentials} and scenario variation. As autonomous vehicles come to be deployed in dynamic environments and situations requiring tight interaction with other intelligent agents, we find that they can push the limits and assumptions that normally underpin classical methods of policy synthesis, such as trajectory optimisation- or search-based approaches. In current practice, this challenge is being attacked from many different perspectives, ranging from the addition of special case treatment and  switching logic within a modular sense-plan-act architecture~\cite{yurtsever2020survey, gonzalez2015review} to adopting fully end-to-end learning architectures~\cite{bojarski2016end}. Our philosophy is to seek a balance between the use of state-of-the-art learning-enabled components along with architectures that allow for safety considerations to be built in by-design. 

This approach reflects the subtlety of the currently existing specifications of the task of driving and the gaps inherent in such specifications, which were originally designed for human interpretation. Firstly, there are often exceptions to stated rules, context-dependent conflicting rules,
and guidance of an ‘open nature’, all requiring interpretation
in context. For instance,  Rule 163 of the UK Highway Code~\cite{highway_code} states that after starting
 an overtaking manoeuvre one should \textit{``move back to the
left\footnote{The United Kingdom adopts a left-hand traffic system.} as soon as you can but do not cut in''}. A more explicit specification of driving conduct (e.g. Rule 163) to something more machine interpretable would allow for policy synthesis in the ways we envision. Secondly, we have noted that driving in urban environments is an intrinsically interactive activity, involving several
actors whose internal states may be opaque to the automated vehicle.
As an example, the UK Highway Code asks drivers to not
\textit{``pull out into traffic so as to cause another driver to slow down''}. Ensuring {\textit{demonstrable}} adherence to such specifications requires architectures that carefully integrate perception, prediction, planning and control.

Approaches such as Responsibility-Sensitive Safety (RSS)~\cite{rss} address similar concerns by defining weak requirements from first principles, such as from considerations of vehicle physics and basic visibility limits. While this is aligned with our thinking, we seek more expressivity in the requirements and a more direct integration between this and policy synthesis. Therefore, we employ a multi-level policy synthesis architecture and we combine multiple optimisation methods each tuned to different specifications but together achieving the objectives of safety-by-design. Likewise, even within just the prediction modules, our architectures are designed both to be interpretable to the design and testing teams, as well as amenable to verification with respect to stated run-time performance requirements. Lastly, making such structure explicit allows for the possibility of incorporating other interaction mechanisms, such as game theoretic schemes, to model interactive phenomena, e.g. altruistic yielding in tight merging scenarios~\cite{GearyRG21}.

\subsection{Evaluating safety at the system level}
Safety is sometimes viewed within the Artificial Intelligence community as only a matter of achieving high reliability (e.g., overcoming adversarial examples). However, as Nancy Leveson and others have argued~\cite{leveson2016engineering}, reliability is neither the only necessary condition nor a sufficient condition for safety. Instead, one needs a system-theoretic approach to understanding hazards and overcoming issues arising in the `outer-loops'. In practice, when one considers the system components being discussed above, this calls for scenario-based designs and evaluations in the context of careful analysis of Operational Design Domains (ODDs) and their characteristics. So, going beyond mere risk analysis of individual components, we seek to understand system-level behaviours emerging from realistic forms of component-level errors and performance degradation, and the interaction thereof.

Performing this analysis through extensive real-world testing is a seemingly impossible task due to the enormous size of the space defined by variables that needs to be covered in testing, the long tail of rare events that are the most crucial to be analysed, and the cost attached to all of that even for a limited ODD. Moreover, it is becoming increasingly accepted that disengagement statistics alone may be a poor descriptor for system safety and trustworthiness~\cite{KALRA2016182}. This calls for employing -- as a crucial part of the development process --  large-scale high-fidelity simulation in which the most important variations of a scenario could be synthesised for deeper analysis.

The scale at which high-fidelity simulation needs to be employed could make traditional approaches difficult and computationally costly: simulating the world in sufficient detail (e.g. to capture all the visual light-reflective and radar-reflective properties) followed by accurate simulation of all the sensor characteristics (e.g., RGB cameras, LIDAR, RADAR). On the other hand, the alternative of creating a simplified simulation environment specifically for testing prediction and planning at the level of objects on a map fails to capture the subtlety of the real world and the physical platform. Thus, we explore Perception Error Models (PEMs) that allow for simulations to be well calibrated to the real world conditions in which the system will subsequently be validated without the need to simulate at the level of light transport and sensor physics, hence focusing attention of the entire safety analysis process on the relevant parts of the design space which may be the most sensitive.

This speaks to how one evaluates the system as a whole. On the one hand, simulation-based testing would always be limited by issues of realism, while other formal approaches to verification and validation can suffer from scaling issues in open environments. So, developing a system-level methodology for large-scale simulation-based development work flow is a crucial area of ongoing development.

\subsection{Paper organisation}
The structure of the remainder of this paper is to summarise a few representative published works from Five, focusing attention specifically on motion planning and prediction issues and their overlap with perception models, mainly because it is here that the multi-agent nature of the problem becomes most salient. An exciting area of research not covered here is to further investigate advances in multi-agent aspects of other
 autonomous driving 
problems, e.g. active perception especially when informed by interaction, federated learning and \textit{ad hoc} interaction at a higher level. 

The rest of this paper is organised as follows:
\begin{itemize}
    \item In Section~\ref{planning}, we describe a general approach to imposing safety specifications into planning, and propose an efficient implementation based on this. In particular, we discuss the following two papers:
    \begin{itemize}
        \item Eiras et al., \textbf{A Two-Stage Optimization-Based Motion Planner for Safe Urban Driving}, \textit{IEEE
Transactions on Robotics 38(2)}, 2022.
        \item Pulver et al., \textbf{PILOT: Efficient Planning by Imitation Learning and
Optimisation for Safe Autonomous Driving}, \textit{IEEE/RSJ International Conference on Intelligent Robots and Systems (IROS)}, 2021.
    \end{itemize}
    \item In Section~\ref{prediction} we describe methods for interpretable motion prediction. In particular, we discuss the following papers:
    \begin{itemize}
        \item Albrecht et al., \textbf{Interpretable goal-based prediction and
planning for autonomous driving}, \textit{IEEE International Conference on Robotics and Automation (ICRA)}, 2021.
        \item Hanna et al., \textbf{Interpretable goal recognition in
the presence of occluded factors for autonomous vehicles}, \textit{IEEE/RSJ International Conference on Intelligent Robots and Systems
(IROS)}, 2021.
        \item Brewitt et al., \textbf{GRIT: Fast, Interpretable, and Verifiable Goal Recognition with Learned Decision
Trees for Autonomous Driving}, \textit{IEEE/RSJ International Conference on Intelligent Robots and Systems (IROS)}, 2021.
        \item Gyevnar et al., \textbf{A Human-Centric Method for Generating Causal Explanations in Natural Language for Autonomous Vehicle Motion Planning}, \textit{IJCAI Workshop on Artificial Intelligence for Autonomous Driving}, 2022.
        \item Antonello et al., \textbf{Flash: Fast and Light Motion Prediction for Autonomous Driving with Bayesian Inverse Planning and Learned Motion Profiles}, \textit{IEEE/RSJ International Conference on Intelligent Robots and Systems (IROS)}, 2022.
    \end{itemize}
    \item In Section~\ref{PEMs} we describe how we use perception error models to inform large-scale simulations and the paper: Sadeghi et al., \textbf{A Step Towards Efficient Evaluation of Complex
Perception Tasks in Simulation}, \textit{NeurIPS Workshop on Machine Learning for Autonomous Driving}, 2021.
\end{itemize}

\section{Components for System-level Safety  at Five}
\subsection{Safe planning\label{planning}}
    Instead of attempting to post-validate the safety of a planning system that does not have safety built-in, e.g. a manually-tuned rule-based system, a better approach is to directly consider safety specifications at design time, such as by employing a constrained optimisation scheme with safety constraints. For autonomous driving, safe and assertive driving style could be captured with \textit{hard constraints} that delimit the domain in which driving is feasible (e.g. kinematic feasibility) and safe (e.g. avoiding the road boundaries or colliding with other road users); and \textit{soft constraints} that capture desirable qualities of  driving (e.g. continuous progress to the target, and having low jerk in both acceleration and steering). Moreover, different contexts suggest additional constraints that capture the subtlety of interaction and expected behaviour in these particular contexts.

With nonlinear dynamics and non-convex constraints, a natural formulation of this planning problem is with constrained Nonlinear Programming (NLP), usually implemented with a receding horizon. Here, hard constraints get introduced into the optimisation objective function through Lagrange multiplier terms (for equality constraints) and KKT conditions (for inequality constraints), and the gradient of this augmented objective function is relied upon to lead to convergence to a solution. However, it is well known for this kind of formalisation that convergence is uncertain and sensitive to initialisation, and solutions tend to be only locally-optimal~\cite{fast_nonlin}.

\subsubsection{Improving convergence rate and solution quality}
\begin{figure}[h]
\centering
\includegraphics[width=9cm]{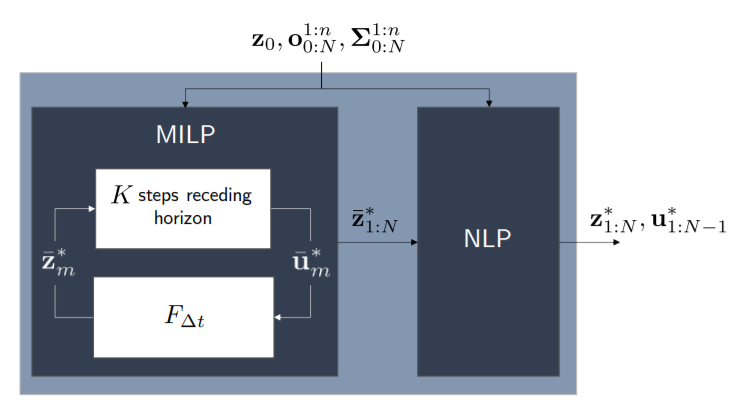}
\caption{2s-OPT architecture. The first stage solves a Mixed-Integer Linear Programming formulation of the planning problem with receding horizon, which solution warm-starts a Nonlinear Programming solver. \copyright [2022] IEEE. Reprinted, with permission, from~\cite{obp}.}
\label{obp}
\end{figure}

In~\cite{obp} we warm-start the NLP problem by the solution of an ‘easier’ approximation of the planning problem formulated as a Mixed-Integer Linear Programming (MILP) problem with receding horizon. This is  achieved by linearising and simplifying the dynamics and constraints, reformulating the objective function, then solving (almost) optimally using Branch and Bound. The solution of this MILP problem is then used as a starting point for the NLP solver. This initialisation stage along with a one-shot NLP stage leads to a two-stage optimisation framework (2s-OPT) as shown in Figure~\ref{obp} that results in improved convergence, faster NLP solving times, and higher solution quality~\cite{obp}.

This multi-level optimisation scheme also allows us to address complex road rules that admit formal description, such as in Signal Temporal Logic (STL). As described in~\cite{raman2014model}, the satisfaction of STL formulae can be encoded as constraints in a MILP formulation, allowing frameworks such as the above to accommodate encodings of road rules and  specifications from the highway code.

\subsubsection{Improving runtime efficiency}
While our two-stage optimisation framework produces high quality solutions, it suffers from the non-trivial additional time required by the solver of the first optimisation stage to converge. This trade-off between quality and runtime efficiency could be accommodated in certain environments but would bar the applicability of the framework to high-speed, interactive settings.

To address this issue, we `distil’ the performance of the 2s-OPT planner offline with a deep neural network, such that the network learns to produce similar outputs to the two-stage optimiser for the same input, and augment that with a light-weight runtime optimisation stage, resulting in the PILOT framework~\cite{pilot}. To that end, we use imitation learning where 2s-OPT acts as the expert. We employ DAgger~\cite{dagger} as a data augmentation scheme and run the learning loop using the CARLA simulator~\cite{Dosovitskiy17}. The neural network has convolutional layers that consume visual representations of the driving scene (and how it is predicted to evolve over the planning horizon), in addition to feed-forward layers that take into account other scalar quantities. The desired output of the network is a smooth, feasible trajectory for the full planning horizon.

\begin{figure}[h]
\centering
\includegraphics[width=10cm]{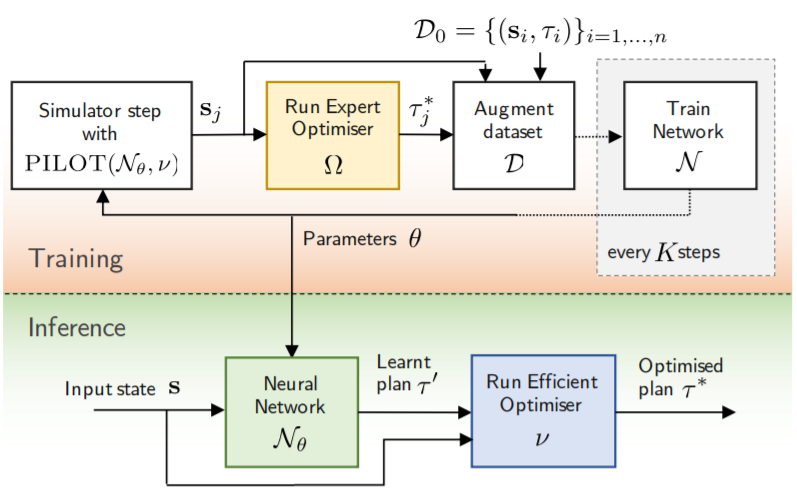}
\caption{PILOT training and runtime pipelines. At the training phase, 2s-OPT is used as the expert to train an imitation deep neural network. At runtime, the output of the network is a seed to a Nonlinear Programming optimisation stage that rectifies the network's output and establishes some of the guarantees. \copyright [2021] IEEE. Reprinted, with permission, from~\cite{pilot}.}
\label{pilot}
\end{figure}

At runtime, in order to retain some of the guarantees with respect to specifications,  we use the output of the network to warm-start an efficient NLP stage. While this hybrid solution would not guarantee the satisfaction of all the specification at all times (that is, if the NLP stage fails to converge within the allocated time budget as discussed earlier), the quality of the initialisation produced by the deep network along with the high-frequency planning allowed by the improved runtime performance  lead to very encouraging empirical results~\cite{pilot}. Figure~\ref{pilot} shows the training and runtime pipelines of PILOT.

Further performance gains could be attained at the neural network training phase through adversarial techniques, e.g. using robust adversarial training~\cite{parot}.

\subsection{Interpretable prediction and its integration with planning systems\label{prediction}}
While interpretability by itself is not a sufficient condition  for safe prediction systems, it is a necessary requirement and a stepping stone towards a less opaque and more introspective stack. Thus, we developed a number of interpretable, model-based prediction systems which we describe next.

We developed an integrated goal recognition and motion planning system, called \textit{Interpretable Goal-based Prediction and Planning} (IGP2)~\cite{igp2}. IGP2 infers posterior probabilities of possible targets extracted from a map and corresponding trajectories of other vehicles using rational inverse planning. The goal and trajectory predictions inform a sampling-based motion planner employing Monte Carlo Tree Search (MCTS), which repeatedly forward-simulates the present states to determine the best sequence of actions for the ego vehicle. By modelling the actions of all vehicles using high-level manoeuvres and macro actions, IGP2 generates manoeuvre plans over extended horizons for which we can extract intuitive explanations based on rationality principles. We tested IGP2 in diverse simulated urban driving scenarios, demonstrating its ability to robustly discover goals and driving intentions of nearby vehicles and exploit this information to generate efficient driving plans for the ego vehicle~\cite{igp2}. An open-source implementation of IGP2 has been released.\footnote{The released implementation, developed in collaboration with the University of Edinburgh, is compatible with the OpenDrive format and integrated with the CARLA open simulation platform~\cite{Dosovitskiy17},  and is available at \url{https://github.com/uoe-agents/IGP2}}

We extended IGP2 by adding the ability to infer the presence of occluded objects. Our approach, \textit{Goal and Occluded Factor Inference} (GOFI)~\cite{gofi}, jointly models the probability of existence of occluded objects along with the goals of other vehicles. If an observed vehicle's behaviour seems rational under the hypothesis of an occluded object being present but seems irrational otherwise, GOFI uses this information to increase its belief that an occluded object is present. Like IGP2, GOFI uses the inferred goal and occluded object probabilities to inform an MCTS planning procedure. In a range of simulated driving scenarios, we showed that GOFI is able to reduce the number of collisions with occluded objects or other vehicles~\cite{gofi}.
    
Towards verifiable prediction for autonomous vehicles, we developed a goal recognition method called \textit{Goal Recognition with Interpretable Trees} (GRIT)~\cite{grit}. Our agenda in this work was to develop a method that is fast, accurate, interpretable, and formally verifiable against specified properties. Unlike previous methods, GRIT is the first method which satisfies all four of these objectives. GRIT infers posterior probabilities over the goals of vehicles using decision trees which are learned from vehicle-trajectory data. We evaluated GRIT across four urban driving scenarios from two vehicle-trajectory data sets. Our experiments showed that GRIT is computationally fast and achieved similar goal recognition accuracy to deep learning based methods and higher accuracy than several other baselines~\cite{grit}. We also showed that GRIT produces trees which are human-interpretable and which can be formally verified against specified properties  by mapping the trees into propositional logic and applying Satisfiability Modulo Theory (SMT) solvers~\cite{grit}.

Based on our goal recognition and motion planning system IGP2, we ask whether we can generate intuitive dialogue-oriented explanations automatically in a way that builds \textit{transparency} and \textit{trust} in our system. IGP2 is particularly well-suited for this task as it was designed from the start with interpretability in mind. We are working towards building an explanation generation system on top of IGP2 that can automatically create intuitive explanations of vehicle decisions in response to natural language explanation queries by human users. A recent prototype and results were presented in~\cite{gyevnar2022humancentric}.

Finally, considering the requirements of a safety-critical and real-time motion planning system, we developed a modular, light-weight motion prediction system called Flash~\cite{flash}. Building on top of inverse planning and goal recognition, Flash has a hybrid architecture that involves both analytically-modelled and data-driven components. This system is designed to exhibit multiple properties of interest: multi-modality, predicted spatial uncertainty, accuracy and computational efficiency. In contrast to many end-to-end regression-based systems, Flash's control-based trajectory generator provides guarantees that kinematic and motion profile constraints are respected, and its modularity facilitates component-focused improvements. In~\cite{flash}, we performed a detailed analysis of the system components with experiments stratifying the data based on behaviours, such as change lane versus follow lane. We demonstrated state-of-the-art performance in terms of trajectory error, especially for longer horizons that are crucial for motion planning. We also showed the ability to isolate performance factors by analysing the impact of the motion profile prediction component and by interpreting the goal recognition outputs.

\subsection{Understanding the limitations of perception systems\label{PEMs}}
While prediction and planning systems normally assume the correctness of their input (the output of perception modules),  no real world sensor systems or real world perception systems are able to produce noise-free output. Understanding the limits of perception systems in a systematic way 1) facilitates developing prediction and planning systems that can handle the noisy real world, and can influence the design of the perception system through its impact on the downstream task of planning~\cite{philion2020learning}; and 2) makes simulations more realistic, closing the gap between simulation and reality.

\subsubsection{Perception error models}

For complex, real-world perception tasks, e.g. compute-intensive object detectors, modern perception pipelines often involve large deep learning models and process quite high-dimensional data~\cite{janai2020computer}. Characterising the emergent error behaviour of an autonomous driving system under test that employs such pipelines usually requires large-scale simulation testing, and hence requires computationally-expensive sensor rendering in addition to the cost of running these pipelines, with quality dependant on the realism of the simulator.
   
\begin{figure}[h]
    \centering
    \includegraphics[width=0.9\textwidth]{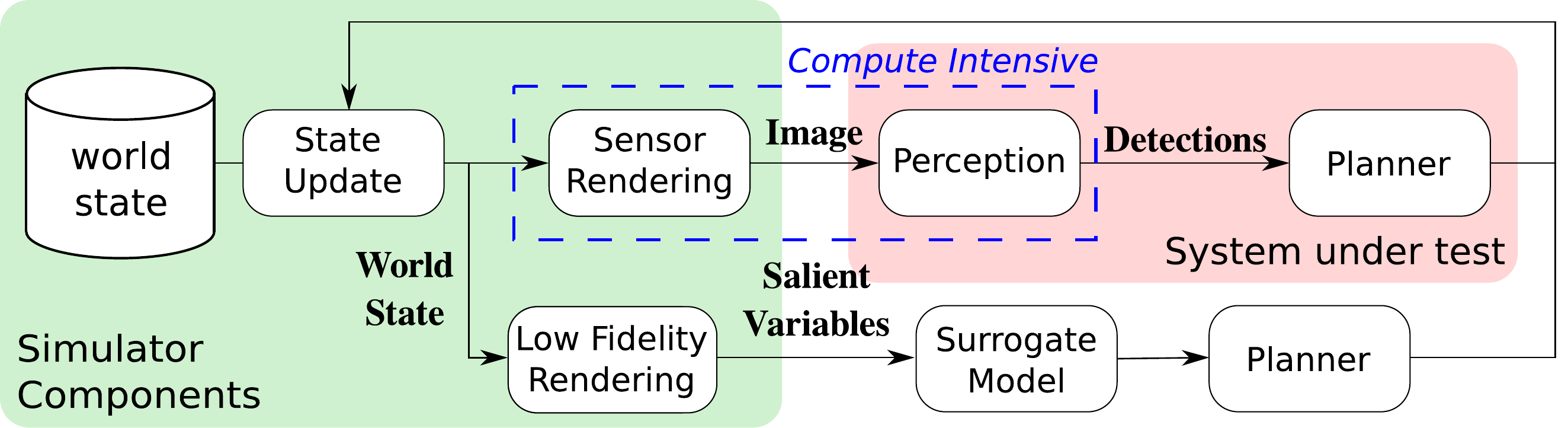}
    \caption{
        Block diagram of a  simulator loop for a backbone task (e.g. LIDAR detector). The compute-intensive stages could be circumvented during end-to-end testing with a surrogate perception error model.
        The outputs of the surrogate model and the backbone task are fed into the model for the downstream task (e.g. a planner) in the same way because their outputs (e.g. detections) are elements of the same vector space.
        The simulator updates the world state based on the output of the downstream task (e.g. the new plan), which is used both by the expensive simulator to produce sensor readings (e.g. LIDAR point clouds) and by the efficient low-fidelity simulator to produce low-dimensional inputs (salient variables) required by the surrogate model. 
    }
    \label{fig:prism_1}
\end{figure}

As an alternative, perception error models (PEMs) serve as a surrogate for the compute-intensive components and pipelines of the task under test, and thereby enable efficient large-scale testing using simplified low-fidelity simulators without the computational cost of executing the expensive deep learning models~\cite{hoss2021review}. Such perception error models could be created using engineering knowledge of the task~\cite{piazzoni2020modeling}, from real world data~\cite{zec2018statistical}, or even using simulated data~\cite{prism}. An example of a simulation pipeline with a PEM is shown in Figure~\ref{fig:prism_1}.

\begin{figure}[h]
     \centering
          \begin{subfigure}[b]{0.3\linewidth}
         \centering
         \includegraphics[width=\linewidth]{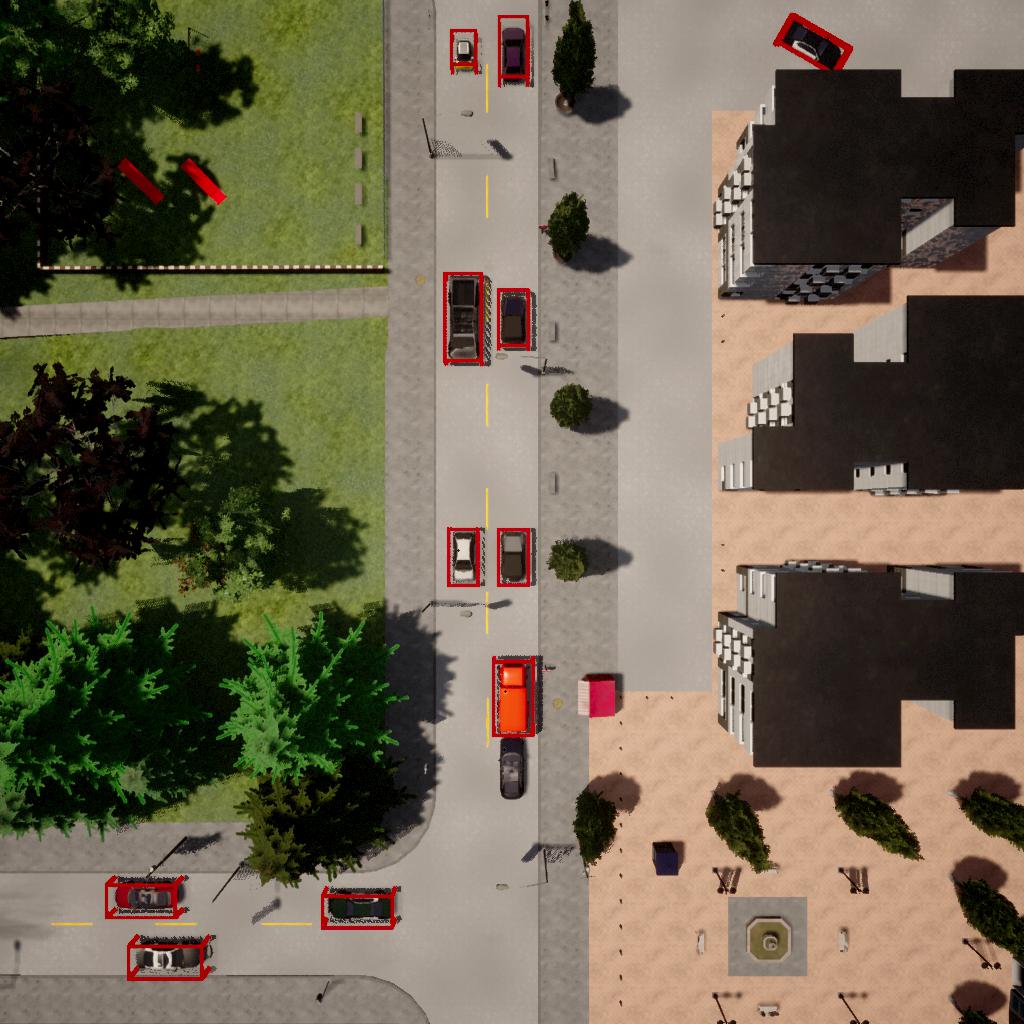}
         \caption{Ground truth}
     \end{subfigure}
     \begin{subfigure}[b]{0.3\linewidth}
         \centering
         \includegraphics[width=\linewidth]{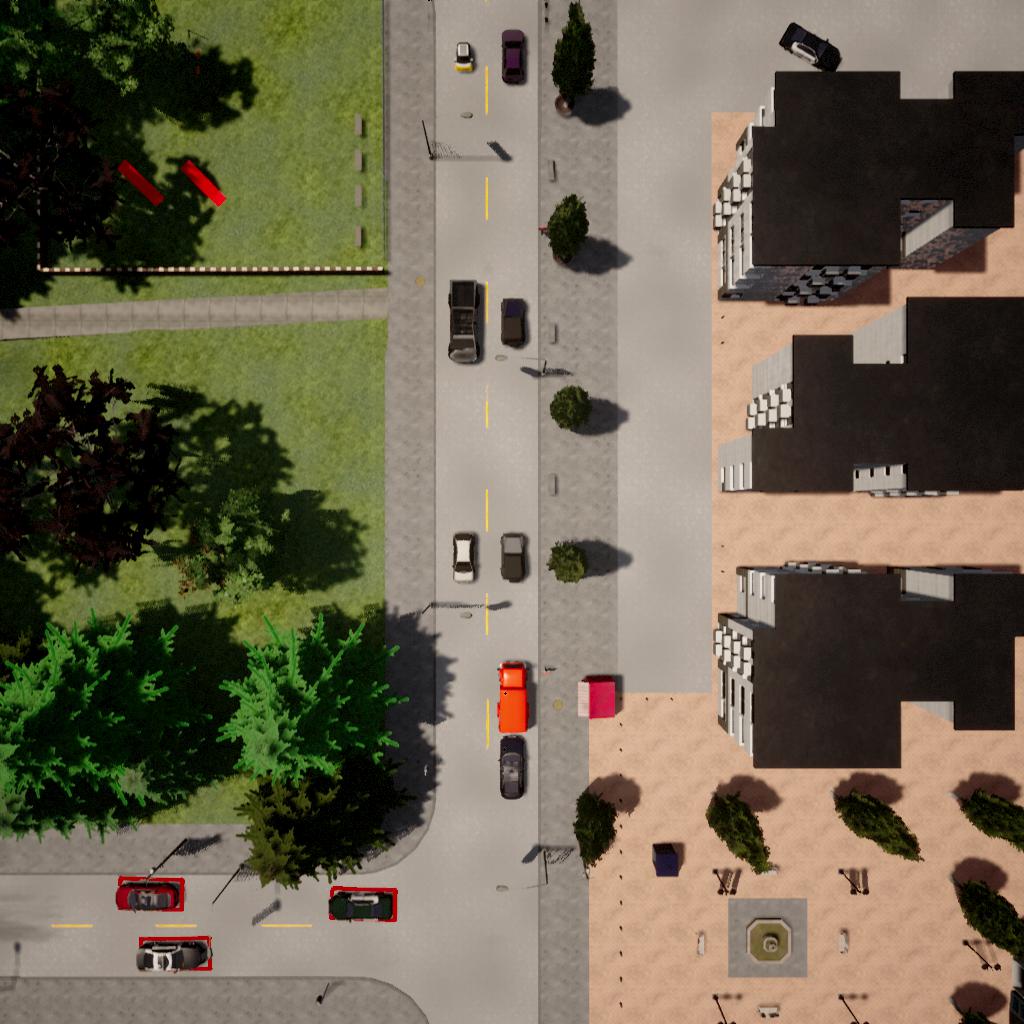}
         \caption{LIDAR detector output}
     \end{subfigure}
      \begin{subfigure}[b]{0.3\linewidth}
         \centering
         \includegraphics[width=\linewidth]{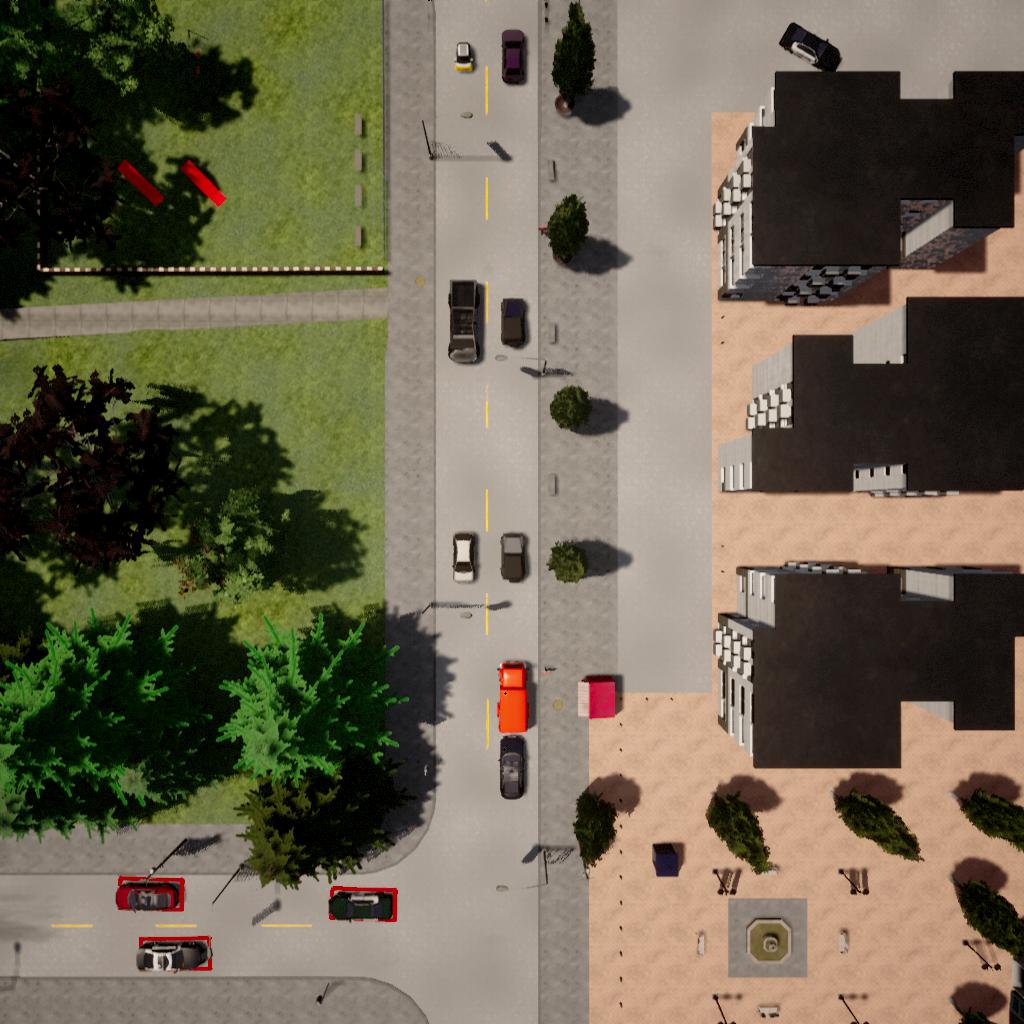}
         \caption{Neural surrogate PEM}
     \end{subfigure}
        \caption{An example of a false negative vehicle detection  that leads to a collision in a CARLA simulation. Ego is the black vehicle in the bottom. Detected objects are shown by red bounding boxes for the ground truth annotation, the LIDAR detector output, and the neural surrogate model output.}
        \label{fig:example_scenes_prism}
\end{figure}

We demonstrate the efficacy of PEMs in~\cite{prism}  through the reduced computational cost of an autonomous driving task in the CARLA simulator~\cite{Dosovitskiy17} when using trained, efficient surrogate models of state-of-the-art LIDAR detectors whilst maintaining the accuracy of the simulation as measured by planning metrics of the behaviour of the ego vehicle and metrics of the performance of the perception system. Figure~\ref{fig:example_scenes_prism} shows an example of using this model where the surrogate PEM model performs similarly to the actual LIDAR detector from the point of view of missing detections.

\section{Conclusion}
We have presented a sample of results taken from published work by the team at Five, addressing aspects of safe-by-design and verifiable autonomous driving in multi-agent interaction scenarios. While these ideas have already been implemented successfully in deployment, sufficient to support field trials in complex urban environments such as in the city of London, we believe that there is much work still to be done. As the ODDs within which autonomous vehicles are deployed become more complex, there will be the need to scale up all components in the development pipeline and deployed stack, ranging from simulations that are extensive enough to capture complex edge cases to data-driven prediction and planning. However, we expect the design principles articulated here to continue to serve as useful guidance also for future systems addressing emerging challenges.

\nocite{*}
\bibliographystyle{ios1}          
\bibliography{bibliography}       
\end{document}

%% file: stack.tikz
\tikzset{every picture/.style={line width=0.75pt}} 

\begin{tikzpicture}[x=0.75pt,y=0.75pt,yscale=-1,xscale=0.97]
\fontfamily{Roboto}

\draw  [color={rgb, 255:red, 0; green, 0; blue, 0 }  ,draw opacity=1 ][fill={rgb, 255:red, 74; green, 144; blue, 226 }  ,fill opacity=1 ] (3.86,34.23) .. controls (3.86,25.82) and (10.68,19) .. (19.09,19) -- (64.77,19) .. controls (73.18,19) and (80,25.82) .. (80,34.23) -- (80,276.77) .. controls (80,285.18) and (73.18,292) .. (64.77,292) -- (19.09,292) .. controls (10.68,292) and (3.86,285.18) .. (3.86,276.77) -- cycle ;
\draw  [color={rgb, 255:red, 0; green, 0; blue, 0 }  ,draw opacity=1 ][fill={rgb, 255:red, 74; green, 144; blue, 226 }  ,fill opacity=1 ] (104.76,49.9) .. controls (104.76,42.38) and (110.86,36.29) .. (118.38,36.29) -- (159.24,36.29) .. controls (166.76,36.29) and (172.86,42.38) .. (172.86,49.9) -- (172.86,133.67) .. controls (172.86,141.19) and (166.76,147.29) .. (159.24,147.29) -- (118.38,147.29) .. controls (110.86,147.29) and (104.76,141.19) .. (104.76,133.67) -- cycle ;
\draw    (81,89.79) -- (103.6,89.79) ;
\draw [shift={(105.6,89.79)}, rotate = 180] [color={rgb, 255:red, 0; green, 0; blue, 0 }  ][line width=0.75]    (10.93,-3.29) .. controls (6.95,-1.4) and (3.31,-0.3) .. (0,0) .. controls (3.31,0.3) and (6.95,1.4) .. (10.93,3.29)   ;
\draw  [color={rgb, 255:red, 0; green, 0; blue, 0 }  ,draw opacity=1 ][fill={rgb, 255:red, 74; green, 144; blue, 226 }  ,fill opacity=1 ] (103,188.26) .. controls (103,180.54) and (109.26,174.29) .. (116.97,174.29) -- (158.89,174.29) .. controls (166.6,174.29) and (172.86,180.54) .. (172.86,188.26) -- (172.86,259.31) .. controls (172.86,267.03) and (166.6,273.29) .. (158.89,273.29) -- (116.97,273.29) .. controls (109.26,273.29) and (103,267.03) .. (103,259.31) -- cycle ;
\draw    (81,225.29) -- (100.86,225.29) ;
\draw [shift={(102.86,225.29)}, rotate = 180] [color={rgb, 255:red, 0; green, 0; blue, 0 }  ][line width=0.75]    (10.93,-3.29) .. controls (6.95,-1.4) and (3.31,-0.3) .. (0,0) .. controls (3.31,0.3) and (6.95,1.4) .. (10.93,3.29)   ;
\draw  [fill={rgb, 255:red, 65; green, 117; blue, 5 }  ,fill opacity=1 ] (199,70.66) .. controls (199,64.22) and (204.22,59) .. (210.66,59) -- (267.2,59) .. controls (273.64,59) and (278.86,64.22) .. (278.86,70.66) -- (278.86,105.63) .. controls (278.86,112.07) and (273.64,117.29) .. (267.2,117.29) -- (210.66,117.29) .. controls (204.22,117.29) and (199,112.07) .. (199,105.63) -- cycle ;
\draw    (171.86,89.79) -- (195.86,89.79) ;
\draw [shift={(197.86,89.79)}, rotate = 180] [color={rgb, 255:red, 0; green, 0; blue, 0 }  ][line width=0.75]    (10.93,-3.29) .. controls (6.95,-1.4) and (3.31,-0.3) .. (0,0) .. controls (3.31,0.3) and (6.95,1.4) .. (10.93,3.29)   ;
\draw  [fill={rgb, 255:red, 65; green, 117; blue, 5 }  ,fill opacity=1 ] (207,163.66) .. controls (207,157.22) and (212.22,152) .. (218.66,152) -- (270.2,152) .. controls (276.64,152) and (281.86,157.22) .. (281.86,163.66) -- (281.86,198.63) .. controls (281.86,205.07) and (276.64,210.29) .. (270.2,210.29) -- (218.66,210.29) .. controls (212.22,210.29) and (207,205.07) .. (207,198.63) -- cycle ;
\draw    (173.86,226.29) -- (204.38,198.63) ;
\draw [shift={(205.86,197.29)}, rotate = 137.82] [color={rgb, 255:red, 0; green, 0; blue, 0 }  ][line width=0.75]    (10.93,-3.29) .. controls (6.95,-1.4) and (3.31,-0.3) .. (0,0) .. controls (3.31,0.3) and (6.95,1.4) .. (10.93,3.29)   ;
\draw  [fill={rgb, 255:red, 65; green, 117; blue, 5 }  ,fill opacity=1 ] (206,231.66) .. controls (206,225.22) and (211.22,220) .. (217.66,220) -- (269.2,220) .. controls (275.64,220) and (280.86,225.22) .. (280.86,231.66) -- (280.86,266.63) .. controls (280.86,273.07) and (275.64,278.29) .. (269.2,278.29) -- (217.66,278.29) .. controls (211.22,278.29) and (206,273.07) .. (206,266.63) -- cycle ;
\draw    (173.86,226.29) -- (204.16,245.23) ;
\draw [shift={(205.86,246.29)}, rotate = 212.01] [color={rgb, 255:red, 0; green, 0; blue, 0 }  ][line width=0.75]    (10.93,-3.29) .. controls (6.95,-1.4) and (3.31,-0.3) .. (0,0) .. controls (3.31,0.3) and (6.95,1.4) .. (10.93,3.29)   ;
\draw  [fill={rgb, 255:red, 65; green, 117; blue, 5 }  ,fill opacity=1 ] (341,167.77) .. controls (341,162.93) and (344.93,159) .. (349.77,159) -- (376.09,159) .. controls (380.93,159) and (384.86,162.93) .. (384.86,167.77) -- (384.86,230.51) .. controls (384.86,235.36) and (380.93,239.29) .. (376.09,239.29) -- (349.77,239.29) .. controls (344.93,239.29) and (341,235.36) .. (341,230.51) -- cycle ;
\draw    (342.86,119.29) -- (364.88,158.54) ;
\draw [shift={(365.86,160.29)}, rotate = 240.71] [color={rgb, 255:red, 0; green, 0; blue, 0 }  ][line width=0.75]    (10.93,-3.29) .. controls (6.95,-1.4) and (3.31,-0.3) .. (0,0) .. controls (3.31,0.3) and (6.95,1.4) .. (10.93,3.29)   ;
\draw  [fill={rgb, 255:red, 65; green, 117; blue, 5 }  ,fill opacity=1 ] (301,72.66) .. controls (301,66.22) and (306.22,61) .. (312.66,61) -- (351.2,61) .. controls (357.64,61) and (362.86,66.22) .. (362.86,72.66) -- (362.86,107.63) .. controls (362.86,114.07) and (357.64,119.29) .. (351.2,119.29) -- (312.66,119.29) .. controls (306.22,119.29) and (301,114.07) .. (301,107.63) -- cycle ;
\draw    (141.86,175.29) -- (141.86,148.29) ;
\draw [shift={(141.86,146.29)}, rotate = 90] [color={rgb, 255:red, 0; green, 0; blue, 0 }  ][line width=0.75]    (10.93,-3.29) .. controls (6.95,-1.4) and (3.31,-0.3) .. (0,0) .. controls (3.31,0.3) and (6.95,1.4) .. (10.93,3.29)   ;
\draw    (241.86,151.29) -- (241.99,118) ;
\draw [shift={(242,116)}, rotate = 90.23] [color={rgb, 255:red, 0; green, 0; blue, 0 }  ][line width=0.75]    (10.93,-3.29) .. controls (6.95,-1.4) and (3.31,-0.3) .. (0,0) .. controls (3.31,0.3) and (6.95,1.4) .. (10.93,3.29)   ;
\draw    (328.86,119.29) -- (283.31,162.28) ;
\draw [shift={(281.86,163.66)}, rotate = 316.65] [color={rgb, 255:red, 0; green, 0; blue, 0 }  ][line width=0.75]    (10.93,-3.29) .. controls (6.95,-1.4) and (3.31,-0.3) .. (0,0) .. controls (3.31,0.3) and (6.95,1.4) .. (10.93,3.29)   ;
\draw    (172.86,123.67) -- (416.4,123.67) ;
\draw [shift={(418.4,123.67)}, rotate = 180] [color={rgb, 255:red, 0; green, 0; blue, 0 }  ][line width=0.75]    (10.93,-3.29) .. controls (6.95,-1.4) and (3.31,-0.3) .. (0,0) .. controls (3.31,0.3) and (6.95,1.4) .. (10.93,3.29)   ;
\draw    (282.8,178.2) -- (338.96,196.66) ;
\draw [shift={(340.86,197.29)}, rotate = 198.2] [color={rgb, 255:red, 0; green, 0; blue, 0 }  ][line width=0.75]    (10.93,-3.29) .. controls (6.95,-1.4) and (3.31,-0.3) .. (0,0) .. controls (3.31,0.3) and (6.95,1.4) .. (10.93,3.29)   ;
\draw    (281.6,249) -- (339.35,198.6) ;
\draw [shift={(340.86,197.29)}, rotate = 138.89] [color={rgb, 255:red, 0; green, 0; blue, 0 }  ][line width=0.75]    (10.93,-3.29) .. controls (6.95,-1.4) and (3.31,-0.3) .. (0,0) .. controls (3.31,0.3) and (6.95,1.4) .. (10.93,3.29)   ;
\draw    (172.86,132.67) -- (349.79,158.71) ;
\draw [shift={(351.77,159)}, rotate = 188.37] [color={rgb, 255:red, 0; green, 0; blue, 0 }  ][line width=0.75]    (10.93,-3.29) .. controls (6.95,-1.4) and (3.31,-0.3) .. (0,0) .. controls (3.31,0.3) and (6.95,1.4) .. (10.93,3.29)   ;
\draw  [fill={rgb, 255:red, 108; green, 172; blue, 46 }  ,fill opacity=0.21 ] (93.86,35) .. controls (93.86,25.61) and (101.47,18) .. (110.86,18) -- (375.86,18) .. controls (385.25,18) and (392.86,25.61) .. (392.86,35) -- (392.86,274) .. controls (392.86,283.39) and (385.25,291) .. (375.86,291) -- (110.86,291) .. controls (101.47,291) and (93.86,283.39) .. (93.86,274) -- cycle ;
\draw  [fill={rgb, 255:red, 165; green, 51; blue, 10 }  ,fill opacity=0.76 ] (417.76,65.9) .. controls (417.76,62.25) and (420.73,59.29) .. (424.38,59.29) -- (444.24,59.29) .. controls (447.89,59.29) and (450.86,62.25) .. (450.86,65.9) -- (450.86,246.95) .. controls (450.86,250.61) and (447.89,253.57) .. (444.24,253.57) -- (424.38,253.57) .. controls (420.73,253.57) and (417.76,250.61) .. (417.76,246.95) -- cycle ;
\draw    (363.86,89.79) -- (416.8,89.79) ;
\draw [shift={(418.8,89.79)}, rotate = 180] [color={rgb, 255:red, 0; green, 0; blue, 0 }  ][line width=0.75]    (10.93,-3.29) .. controls (6.95,-1.4) and (3.31,-0.3) .. (0,0) .. controls (3.31,0.3) and (6.95,1.4) .. (10.93,3.29)   ;
\draw    (451.86,156.29) -- (466.86,156.29) ;
\draw [shift={(468.86,156.29)}, rotate = 180] [color={rgb, 255:red, 0; green, 0; blue, 0 }  ][line width=0.75]    (10.93,-3.29) .. controls (6.95,-1.4) and (3.31,-0.3) .. (0,0) .. controls (3.31,0.3) and (6.95,1.4) .. (10.93,3.29)   ;
\draw  [fill={rgb, 255:red, 199; green, 60; blue, 60 }  ,fill opacity=0.12 ] (406.86,30.84) .. controls (406.86,23.75) and (412.6,18) .. (419.69,18) -- (600.16,18) .. controls (607.25,18) and (613,23.75) .. (613,30.84) -- (613,278.16) .. controls (613,285.25) and (607.25,291) .. (600.16,291) -- (419.69,291) .. controls (412.6,291) and (406.86,285.25) .. (406.86,278.16) -- cycle ;
\draw  [fill={rgb, 255:red, 165; green, 51; blue, 10 }  ,fill opacity=0.76 ] (468.76,65.9) .. controls (468.76,62.25) and (471.73,59.29) .. (475.38,59.29) -- (495.24,59.29) .. controls (498.89,59.29) and (501.86,62.25) .. (501.86,65.9) -- (501.86,244.95) .. controls (501.86,248.61) and (498.89,251.57) .. (495.24,251.57) -- (475.38,251.57) .. controls (471.73,251.57) and (468.76,248.61) .. (468.76,244.95) -- cycle ;
\draw  [fill={rgb, 255:red, 165; green, 51; blue, 10 }  ,fill opacity=0.76 ] (518.76,65.9) .. controls (518.76,62.25) and (521.73,59.29) .. (525.38,59.29) -- (545.24,59.29) .. controls (548.89,59.29) and (551.86,62.25) .. (551.86,65.9) -- (551.86,244.95) .. controls (551.86,248.61) and (548.89,251.57) .. (545.24,251.57) -- (525.38,251.57) .. controls (521.73,251.57) and (518.76,248.61) .. (518.76,244.95) -- cycle ;
\draw  [fill={rgb, 255:red, 219; green, 142; blue, 65 }  ,fill opacity=1 ] (570.76,65.9) .. controls (570.76,62.25) and (573.73,59.29) .. (577.38,59.29) -- (597.24,59.29) .. controls (600.89,59.29) and (603.86,62.25) .. (603.86,65.9) -- (603.86,244.95) .. controls (603.86,248.61) and (600.89,251.57) .. (597.24,251.57) -- (577.38,251.57) .. controls (573.73,251.57) and (570.76,248.61) .. (570.76,244.95) -- cycle ;
\draw    (502.86,156.29) -- (517.86,156.29) ;
\draw [shift={(519.86,156.29)}, rotate = 180] [color={rgb, 255:red, 0; green, 0; blue, 0 }  ][line width=0.75]    (10.93,-3.29) .. controls (6.95,-1.4) and (3.31,-0.3) .. (0,0) .. controls (3.31,0.3) and (6.95,1.4) .. (10.93,3.29)   ;
\draw    (553.86,156.29) -- (568.86,156.29) ;
\draw [shift={(570.86,156.29)}, rotate = 180] [color={rgb, 255:red, 0; green, 0; blue, 0 }  ][line width=0.75]    (10.93,-3.29) .. controls (6.95,-1.4) and (3.31,-0.3) .. (0,0) .. controls (3.31,0.3) and (6.95,1.4) .. (10.93,3.29)   ;
\draw  [fill={rgb, 255:red, 165; green, 51; blue, 10 }  ,fill opacity=0.76 ] (450,266.46) .. controls (450,264.55) and (451.55,263) .. (453.46,263) -- (579.54,263) .. controls (581.45,263) and (583,264.55) .. (583,266.46) -- (583,276.83) .. controls (583,278.74) and (581.45,280.29) .. (579.54,280.29) -- (453.46,280.29) .. controls (451.55,280.29) and (450,278.74) .. (450,276.83) -- cycle ;
\draw  [color={rgb, 255:red, 0; green, 0; blue, 0 }  ,draw opacity=1 ][fill={rgb, 255:red, 74; green, 144; blue, 226 }  ,fill opacity=1 ] (624.86,24.83) .. controls (624.86,21.61) and (627.47,19) .. (630.69,19) -- (648.17,19) .. controls (651.39,19) and (654,21.61) .. (654,24.83) -- (654,286.17) .. controls (654,289.39) and (651.39,292) .. (648.17,292) -- (630.69,292) .. controls (627.47,292) and (624.86,289.39) .. (624.86,286.17) -- cycle ;
\draw    (605.03,157.5) -- (622.8,157.41) ;
\draw [shift={(624.8,157.4)}, rotate = 179.71] [color={rgb, 255:red, 0; green, 0; blue, 0 }  ][line width=0.75]    (10.93,-3.29) .. controls (6.95,-1.4) and (3.31,-0.3) .. (0,0) .. controls (3.31,0.3) and (6.95,1.4) .. (10.93,3.29)   ;
\draw  [fill={rgb, 255:red, 65; green, 117; blue, 5 }  ,fill opacity=1 ] (3.86,309.14) .. controls (3.86,303.07) and (8.78,298.14) .. (14.86,298.14) -- (644.86,298.14) .. controls (650.93,298.14) and (655.86,303.07) .. (655.86,309.14) -- (655.86,309.14) .. controls (655.86,315.22) and (650.93,320.14) .. (644.86,320.14) -- (14.86,320.14) .. controls (8.78,320.14) and (3.86,315.22) .. (3.86,309.14) -- cycle ;
\draw  [fill={rgb, 255:red, 65; green, 117; blue, 5 }  ,fill opacity=1 ] (3.86,338.14) .. controls (3.86,332.07) and (8.78,327.14) .. (14.86,327.14) -- (645,327.14) .. controls (651.08,327.14) and (656,332.07) .. (656,338.14) -- (656,338.14) .. controls (656,344.22) and (651.08,349.14) .. (645,349.14) -- (14.86,349.14) .. controls (8.78,349.14) and (3.86,344.22) .. (3.86,338.14) -- cycle ;
\draw    (384.49,197.94) -- (416.09,197.94) ;
\draw [shift={(418.09,197.94)}, rotate = 180] [color={rgb, 255:red, 0; green, 0; blue, 0 }  ][line width=0.75]    (10.93,-3.29) .. controls (6.95,-1.4) and (3.31,-0.3) .. (0,0) .. controls (3.31,0.3) and (6.95,1.4) .. (10.93,3.29)   ;
\draw    (427.69,17.83) .. controls (408.29,-4.68) and (392.65,2.18) .. (377.28,16.64) ;
\draw [shift={(375.86,18)}, rotate = 315.84] [color={rgb, 255:red, 0; green, 0; blue, 0 }  ][line width=0.75]    (10.93,-3.29) .. controls (6.95,-1.4) and (3.31,-0.3) .. (0,0) .. controls (3.31,0.3) and (6.95,1.4) .. (10.93,3.29)   ;

\draw (41.25,115.21) node  [font=\small,color={rgb, 255:red, 255; green, 255; blue, 255 }  ,opacity=1 ] [align=left] {\textbf{Sensor}\\\textbf{Drivers}\\\textbf{\&}\\\textbf{Callibration}};
\draw (244.43,181.14) node  [font=\small,color={rgb, 255:red, 255; green, 255; blue, 255 }  ,opacity=1 ] [align=left] {\textbf{Static}\\\textbf{Semantics}};
\draw (246.37,247.36) node  [font=\small,color={rgb, 255:red, 255; green, 255; blue, 255 }  ,opacity=1 ] [align=left] {\textbf{Dynamic}\\\textbf{Semantics}};
\draw (364,197.36) node  [font=\small,color={rgb, 255:red, 255; green, 255; blue, 255 }  ,opacity=1 ] [align=left] {\textbf{Fusion}};
\draw (332.38,88.43) node  [font=\small,color={rgb, 255:red, 255; green, 255; blue, 255 }  ,opacity=1 ] [align=left] {\textbf{Mapping}};
\draw (138.62,225.17) node  [font=\small,color={rgb, 255:red, 255; green, 255; blue, 255 }  ,opacity=1 ] [align=left] {\textbf{Geometry}};
\draw (138.35,89.85) node  [font=\small,color={rgb, 255:red, 255; green, 255; blue, 255 }  ,opacity=1 ] [align=left] {\textbf{Odometry}};
\draw (241.13,88) node  [font=\small,color={rgb, 255:red, 255; green, 255; blue, 255 }  ,opacity=1 ] [align=left] {\textbf{Localisation}};
\draw (194,24) node [anchor=north west][inner sep=0.75pt]  [color={rgb, 255:red, 255; green, 255; blue, 255 }  ,opacity=1 ] [align=left] {\textbf{\textcolor[rgb]{0.61,0.61,0.61}{PERCEPTION}}};
\draw (433.85,146.85) node  [font=\small,color={rgb, 255:red, 255; green, 255; blue, 255 }  ,opacity=1 ,rotate=-270] [align=left] {\textbf{Prediction}};
\draw (420,24) node [anchor=north west][inner sep=0.75pt]  [color={rgb, 255:red, 255; green, 255; blue, 255 }  ,opacity=1 ] [align=left] {\textbf{\textcolor[rgb]{0.61,0.61,0.61}{PREDICTION \& PLANNING}}};
\draw (483.85,146.85) node  [font=\small,color={rgb, 255:red, 255; green, 255; blue, 255 }  ,opacity=1 ,rotate=-270] [align=left] {\textbf{Planning}};
\draw (534.85,146.85) node  [font=\small,color={rgb, 255:red, 255; green, 255; blue, 255 }  ,opacity=1 ,rotate=-270] [align=left] {\textbf{Path Synthesis}};
\draw (585.85,146.85) node  [font=\small,color={rgb, 255:red, 255; green, 255; blue, 255 }  ,opacity=1 ,rotate=-270] [align=left] {\textbf{Control}};
\draw (517,271.14) node  [font=\small,color={rgb, 255:red, 255; green, 255; blue, 255 }  ,opacity=1 ] [align=left] {\textbf{Rules of the road}};
\draw (639.85,146.85) node  [font=\small,color={rgb, 255:red, 255; green, 255; blue, 255 }  ,opacity=1 ,rotate=-270] [align=left] {\textbf{Actuator Drivers}};
\draw (6.0,168.5) node [anchor=north west][inner sep=0.75pt]  [color={rgb, 255:red, 255; green, 255; blue, 255 }  ,opacity=1 ] [align=left] {{\tiny Cameras}\\{\tiny Radar}\\{\tiny Lidar }\\{\tiny GPS}\\{\tiny IMU}\\{\tiny Wheel encoders}};
\draw (329.86,308.64) node  [font=\small,color={rgb, 255:red, 255; green, 255; blue, 255 }  ,opacity=1 ] [align=left] {\textbf{Vehicle software infrastructure}};
\draw (331,338.14) node  [font=\small,color={rgb, 255:red, 255; green, 255; blue, 255 }  ,opacity=1 ] [align=left] {\textbf{Safety infrastructure}};

\draw    (280.68,89.96) -- (298.68,89.64) ;
\draw [shift={(300.68,89.6)}, rotate = 178.96] [color={rgb, 255:red, 0; green, 0; blue, 0 }  ][line width=0.75]    (10.93,-4.9) .. controls (6.95,-2.3) and (3.31,-0.67) .. (0,0) .. controls (3.31,0.67) and (6.95,2.3) .. (10.93,4.9)   ;
\draw [shift={(278.68,90)}, rotate = 358.96] [color={rgb, 255:red, 0; green, 0; blue, 0 }  ][line width=0.75]    (10.93,-3.29) .. controls (6.95,-1.4) and (3.31,-0.3) .. (0,0) .. controls (3.31,0.3) and (6.95,1.4) .. (10.93,3.29)   ;

\end{tikzpicture}